\documentclass[lettersize,journal]{IEEEtran}
\usepackage{amsmath,amsfonts}
\usepackage{algorithmic}
\usepackage{algorithm}
\usepackage{array}
\usepackage[caption=false,font=normalsize,labelfont=sf,textfont=sf]{subfig}
\usepackage{textcomp}
\usepackage{stfloats}
\usepackage{url}
\usepackage{verbatim}
\usepackage{graphicx}
\usepackage{cite}
\usepackage{kotex}
\usepackage{color}
\usepackage{hyperref}

\begin{document}

\title{Object-Aware Impedance Control for Human-Robot Collaborative Task with Online Object Parameter Estimation}

\author{Jinseong Park$^{1}$, Yong-Sik Shin$^{1*}$ and Sanghyun Kim$^{2*}$ 
\thanks{*This research was supported by the National Research Council of Science \& Technology as part of the project entitled ``Development of Core Machinery Technologies for Autonomous Operation and Manufacturing'' (NK242G) and the Industrial Strategic Technology Development Program (no. 20018745) funded by the Ministry of Trade, Industry \& Energy (MOTIE, Korea). \textit{(Corresponding author: Yong-Sik Shin and Sanghyun Kim)}}
\thanks{$^{1}$J. Park and Y.-S Shin are senior researchers at the Korea Institute of Machinery \& Materials, Daejeon, South Korea.
{\tt\small jspark2090@kimm.re.kr; yshin86@kimm.re.kr}}%
\thanks{$^{2}$S. Kim is with the Department of Mechanical Engineering, KyungHee University, Yongin, South Korea.
{\tt\small kim87@khu.ac.kr}}%
}

\markboth{}%
{Shell \MakeLowercase{\textit{et al.}}: A Sample Article Using IEEEtran.cls for IEEE Journals}

\IEEEpubid{0000--0000/00\$00.00~\copyright~2023 IEEE}

\maketitle

\begin{abstract}
Physical human-robot interactions (pHRIs) can improve robot autonomy and reduce physical demands on humans. In this paper, we consider a collaborative task with a considerably long object and no prior knowledge of the object's parameters.
An integrated control framework with an online object parameter estimator and a Cartesian object-aware impedance controller is proposed to realize complicated scenarios.
During the transportation task, the object parameters are estimated online while a robot and human lift an object. The perturbation motion is incorporated into the null space of the desired trajectory to enhance the estimator accuracy.
An object-aware impedance controller is designed using the real-time estimation results to effectively transmit the intended human motion to the robot through the object. Experimental demonstrations of collaborative tasks, including object transportation and assembly tasks, are implemented to show the effectiveness of our proposed method.
\end{abstract}

\def\abstractname{Note to Practitioners}
\begin{abstract}
This research was motivated by the need to facilitate collaboration between humans and robots in handling heavy or considerable long objects, which can be challenging for a single operator.
This paper proposes a physical Human-Robot Interaction (pHRI) approach that enables physical interaction between the human and the robot through the object without additional sensors such as camera or human-machine interfaces. 
To achieve collaborative task, the separation between the intended human motion intention and the object dynamics is essential. Most of real-world situations involve uncertain or unknown objects, making it challenging to assume prior knowledge of the target object's properties. Therefore, this research introduces a real-time approach for estimating the dynamic parameters of unknown objects during collaboration, without requiring additional operational time.
Consequently, the design of object-aware impedance controller can be achieved by real-time incorporation of object dynamics. Collaborative transportation and assembly task is demonstrated with whole-body controlled mobile manipulator and 1.5m long object.
In future research, we will focus on enhancing the estimation precision through the use of physical informed neural network methods.
\end{abstract}


\begin{figure}[t]
	\centering
	\includegraphics[clip,width=1.00\columnwidth]{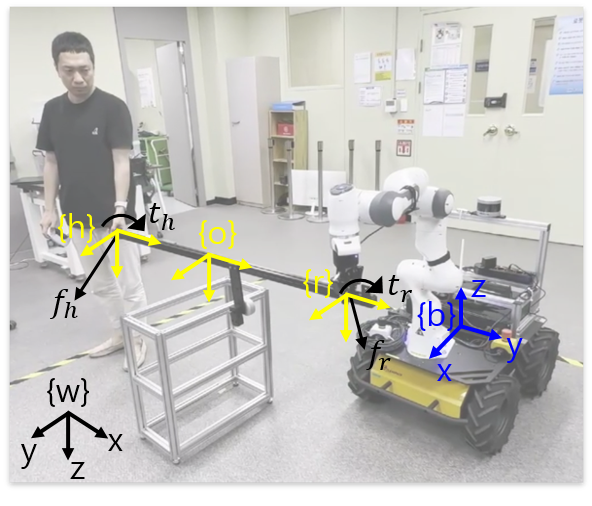}    
\caption{The human-robot collaborative task scenario. The coordinates of the robot and human are represented: \{w\}:world, \{h\}:human, \{o\}:object, \{r\}:end effector of the robot, \{m\}:mobile base.}
	\label{Fig_1}
\end{figure}

\section{Introduction}
\IEEEPARstart{R}{obots} have been increasingly used in human environments for repetitive and simple operations with the notable improvements in the safety and performance of hardware and controllers.
Nevertheless, their applicability is still limited in industrial and service-related scenarios in which flexibility is needed, including manufacturing, logistics, and construction.
On the other hand, robots have several advantages over humans, such as high precision in repetitive tasks and the lack of increasing fatigue during payload lifting.
Therefore, the development of robotic technologies for physical interactions between humans and robots is important to improve the autonomy of robots with human decision-making capabilities and reduce the excessive physical demand on humans during payload lifting \cite{Sirintuna2, Sirintuna, Santis, Mortl, Leonori, Rapetti, Alevizos}.
In physical human-robot interactions (pHRIs), two or more agents are physically coupled and interact and communicate to perform tasks.
pHRIs enable collaborative transportation of heavy or bulky objects and manipulations involving multiple grasping points, which would be difficult to achieve for either humans or robots alone \cite{Cehajic, Torielli}.

\IEEEpubidadjcol

In this paper, we focus on human-robot collaborative transportation and assembly of bulky objects without prior knowledge of the object's parameters. Due to the considerable length of the target object, the human cannot directly contact the robot in the leader-follower collaboration task.
Consequently, the intended human motion can be transmitted to the robot only indirectly through the object.
Additionally, due to the lack of knowledge regarding the object's parameters, an online object parameter estimation process is necessary to accurately determine the intended human motion.

Human-robot collaborative transportation tasks have been studied in various applications \cite{Asfour, Lawitzky, Benzi}.
These tasks are mainly performed under a leader-follower architecture, in which the desired trajectory of the robot is explicitly determined by humans based on the estimated intended human motion.
The effects of dynamic role allocation on physical robotic assistants have been investigated, and the load distribution between the partners was efficiently determined with redundancies of the control inputs \cite{Mortl, Lawitzky}. 
However, only a planar task was performed without considering the inertial effects of the object.
In pHRI-based cooperative object manipulation tasks, the performance estimator-based intention was presented to reduce human effort during collaborative tasks \cite{Alevizos, Alevizos2}. However, the demonstration with a human was performed with direct manipulation of the robot, and the geometry of the object was not considered.
Online adaptation of the admittance control parameters reduced human physical effort during collaborative transportation tasks \cite{Benzi}. Human intention was inferred by monitoring the magnitude and direction of the desired acceleration and the present velocity value. However, the inertial parameters and shapes of the objects have not yet been considered.

The use of external sensors has also been studied for collaborative tasks with physical interactions.
Human-robot collaborative transportation tasks with unknown deformable objects have been performed by utilizing both haptic and kinematic information to estimate the intended human motion \cite{Sirintuna}.
A low-cost cover made of soft, large-area capacitive tactile sensors was placed around the platform base to measure the interaction forces applied to the robot base to implement an expanded Cartesian impedance controller, and follow-me controllers were implemented \cite{Leonori}.
An object lifted by a human and robot was presented using biceps activity, which was measured based on surface electromyography (EMG), to estimate the changes in the user's hand height \cite{Delpreto}.
Invariance control was also presented for safe human-robot interactions in dynamic environments by enforcing dynamic constraints even in the presence of external forces \cite{Kimmel}.
Safe-aware human-robot collaborative transportation tasks considering aerial robots with cable-suspended payloads have also been demonstrated, in which a safe distance is maintained between the human operator and each agent without affecting performance \cite{Liu}.
The use of external sensors enables direct acquisition of motion and wrench information for the human participant, enhancing the performance and stability of collaborative tasks. However, in practice, it may not always be feasible to equip additional sensors in systems.
Therefore, within the context of collaborative tasks, the estimation of intended human motion, even in the absence of such sensors, can be a noteworthy approach, notwithstanding the acceptance of estimation result inaccuracies.

Since the dynamic object parameters are mostly unknown in unstructured environments, online identification strategies for estimating object dynamics should be developed to accurately understand the intended human motion during collaborative tasks.
Bias in the object dynamics leads to inaccurately calculated robot wrenches, which may disturb the human during interactions and affect the trust and interaction behavior of the human partner \cite{Cehajic2, Lawitzky, Mortl}.
An online payload identification approach was presented based on the momentum observer using proprioceptive sensors with a calibration scheme that compensates for static offset between the virtual and real object \cite{Kurdas}. However, directly applying this method in collaborative tasks is not straightforward due to the presence of dynamically coupled human partners.
Alternatively, in some studies, only the weight of the object was estimated during the lifting phase of robot only \cite{Torielli} or human-robot collaborative tasks \cite{Bussy}. However, other dynamic parameters, such as the center of mass and moments of inertia, are also needed depending on the task.
An online grasp pose and object parameter estimation strategy for pHRI was also presented \cite{Cehajic2, Cehajic}. This strategy aims to induce identification-relevant motions that minimally disturb the desired human motion while preventing undesired interaction torques with the human.
The use of probing signals for precise estimation and online estimation strategies is  similar to our objective; however, these strategies require measurements of interaction torques for both the human and the robot and tracking of human motion, which may not always be possible in real-life scenarios.
Indeed, without acquiring the wrench and motion information on the human side, it can be difficult to precisely estimate the object's parameters. However, the objective of this study is to accurately determine the intended human motion while the human and robot are collaboratively lifting the object.
Thus, it is adequate to estimate the effective object parameters, which corresponds to the load endured by the robot during the collaboration task, in this specific configuration.
Consequently, in this study, we estimate the effective object parameters without the use of additional sensors on the human side.
Note that researchers mainly adopted estimator methods based on least squares methods (\cite{Cehajic, Franchi, Kurdas}) or Kalman filter-based methods (\cite{Wuest, Jang} for parameter estimation. We adopted the extended Kalman filter (EKF) because the estimation system varies over time.

In this paper, a Cartesian object-aware impedance controller with online object parameter estimation is proposed. Object parameter estimation enables a precise understanding of the intended human motion during human-robot collaborative tasks.
Furthermore, to effectively transmit the intended human motion to the robot through long objects, the desired inertia shaping of impedance is performed with considering the geometry of the object.

The contributions of this paper can be summarized as follows:

\begin{itemize}
\item The effective object parameters are estimated online by inducing identification motions in the null space of the Cartesian trajectory without the use of additional sensors on the human sides during load-sharing conditions.

\item A Cartesian object-aware impedance controller is designed that effectively transmits the intended human motion to the robot through the object by reflecting the object dynamics to the impedance relations.

\item Experimental demonstrations of a collaborative transportation and the assembly of a long object using a mobile manipulator are presented.
\end{itemize}

The remainder of the paper is organized as follows. In Section II, an online object parameter estimation method is presented by deriving the object dynamic equations. Based on a whole-body control architecture, a Cartesian object-aware impedance controller is designed in Section III. Experimental demonstrations and results are presented in Section IV, and the conclusions are discussed in Section V.

\section{Online Object Parameter Estimation}

\subsection{Object Dynamics}
We investigate a human-robot collaborative task scenario involving an elongated object in which parameters such as the mass, center of mass, and moment of inertia are not provided in advance.

The dynamics of the object are as follows \cite{Cehajic, Kurdas}:
\begin{eqnarray}
	\label{eq_1}
    \begin{array}{l}
    \boldsymbol{M}_o \ddot{\boldsymbol{x}}_o + \boldsymbol{C}_o (\boldsymbol{x}_o, \dot{\boldsymbol{x}}_o ) = \boldsymbol{F}_o
    \end{array}
\end{eqnarray}
where 
$\\$

\hspace{0.5cm}$\boldsymbol{M}_o = \begin{bmatrix} m_o \boldsymbol{I}_3 & \boldsymbol{0}_3 \\ \boldsymbol{0}_3 & \boldsymbol{J}_0 \end{bmatrix} $, 
            $\boldsymbol{C}_o = \begin{bmatrix} -m_o \boldsymbol{g} \\ { \boldsymbol{\omega}}_o \times \boldsymbol{J}_o { \boldsymbol{\omega}}_o \end{bmatrix}$

$\\$
$\boldsymbol{x}_o = [\boldsymbol{p}_o^T , \boldsymbol{q}_o^T]^T \in SE(3)$ is the state of the object expressed in the world frame \{w\}, as illustrated in Fig.~\ref{Fig_1}, and $\boldsymbol{p}_o, \boldsymbol{q}_o$ represent translation and orientation, respectively. Let $\dot{\boldsymbol{x}}_o = [\boldsymbol{v}_o^T , { \boldsymbol{\omega}}_o^T]^T$ and $\ddot{\boldsymbol{x}}_o = [\dot{\boldsymbol{v}}_o^T , \dot{{ \boldsymbol{\omega}}}_o^T]^T$ be twist and acceleration vectors, respectively.
$\boldsymbol{F}_o = [\boldsymbol{f}_o^T , \boldsymbol{t}_o^T]^T \in \mathbb{R}^6$ is the total object wrench, where $\boldsymbol{f}_o \in \mathbb{R}^3$ is the force and $\boldsymbol{t}_o \in \mathbb{R}^3$ is the torque acting on the object. $\boldsymbol{M}_o \in \mathbb{R}^{6 \times 6}$ and $\boldsymbol{C}_o \in \mathbb{R}^6$ are the inertia matrix and nonlinear matrix containing Coriolis effects and gravity, respectively.

Because human and robot wrenches both act on the object, the relative kinematics can be represented as follows:
\begin{eqnarray}
	\label{eq_2}
    \begin{array}{l}
        \boldsymbol{F}_o = \boldsymbol{G}    \begin{bmatrix} \boldsymbol{F}_r \\ \boldsymbol{F}_h \end{bmatrix}     
    \end{array}
\end{eqnarray}
where $\boldsymbol{F}_r = [\boldsymbol{f}_r^T , \boldsymbol{t}_r^T]^T \in \mathbb{R}^6$ and $\boldsymbol{F}_h = [\boldsymbol{f}_h^T , \boldsymbol{t}_h^T]^T \in \mathbb{R}^6$ are the robot and human wrenches at each grasping point, respectively. $\boldsymbol{G}$ represents the kinematic relations in the robot and human frames with respect to the object frame and is defined as follows:
\begin{eqnarray}
	\label{eq_3}
    \begin{array}{l}
        \boldsymbol{G} = \begin{bmatrix} {}^{o}\boldsymbol{G}_r & {}^{o}\boldsymbol{G}_h \end{bmatrix} = 
            \begin{bmatrix} \boldsymbol{I}_3 & \boldsymbol{0}_3 & \boldsymbol{I}_3 & \boldsymbol{0}_3 \\ |{}^{o}\boldsymbol{p}_r|_{\times} & \boldsymbol{I}_3 & |{}^{o}\boldsymbol{p}_h|_{\times} & \boldsymbol{I}_3 \end{bmatrix}
    \end{array}
\end{eqnarray}
where ${}^{o}\boldsymbol{p}_r \in \mathbb{R}^3$ and ${}^{o}\boldsymbol{p}_h \in \mathbb{R}^3$ are vectors from the object to the robot/human grasp points, respectively. $|{}^{o}\boldsymbol{p}_r|_x \in \mathbb{R}^{3 \times 3 }$ denotes the skew matrix, where vector $^{o}\boldsymbol{p}_r$ represents the cross product. $^{o}\boldsymbol{G}_r \in \mathbb{R}^{6 \times 6}$ and ${}^{o}\boldsymbol{G}_h \in \mathbb{R}^{6 \times 6}$ are partial grasp matrices from the object to the robot/human grasp points, respectively.

The kinematic relations between the object frame and robot frame can be represented in terms of the velocity:
\begin{eqnarray}
	\label{eq_4}
    \begin{array}{l}
    \begin{bmatrix} \boldsymbol{v}_o \\ { \boldsymbol{\omega}}_o \end{bmatrix} =
    \begin{bmatrix} \boldsymbol{v}_r + { \boldsymbol{\omega}}_r \times {}^{r}\boldsymbol{p}_o \\{ \boldsymbol{\omega}}_r \end{bmatrix} =
    \begin{bmatrix} \boldsymbol{I}_3 & -|{}^{r}\boldsymbol{p}_o|_{\times} \\ \boldsymbol{0}_3 & \boldsymbol{I}_3 \end{bmatrix}
    \begin{bmatrix} \boldsymbol{v}_r \\ { \boldsymbol{\omega}}_r  \end{bmatrix}     
    \end{array}
\end{eqnarray}

They can also be represented in terms of the acceleration as follows:
\begin{eqnarray}
	\label{eq_5}
    \begin{array}{l}
    \begin{bmatrix} \dot{\boldsymbol{v}_o} \\ \dot{ \boldsymbol{\omega}}_o \end{bmatrix} =
    \begin{bmatrix} \dot{\boldsymbol{v}}_r + \dot{ \boldsymbol{\omega}}_r \times {}^{r}\boldsymbol{p}_o +  \boldsymbol{\omega}_r \times ( \boldsymbol{\omega}_r \times {}^{r}\boldsymbol{p}_o) \\ \dot{ \boldsymbol{\omega}}_r \end{bmatrix} 
    \end{array}
\end{eqnarray}

According to (\ref{eq_2}), $F_o$ can be represented by ${}^r\boldsymbol{F}_o$ as follows:
\begin{eqnarray}
	\label{eq_6}
    \begin{array}{l}
        \boldsymbol{F}_o = {}^{o}\boldsymbol{G}_r \times {{}^{r}\boldsymbol{F}_o}        
    \end{array}
\end{eqnarray}
where

\hspace{1.0cm}${}^o\boldsymbol{G}_r = \begin{bmatrix} \boldsymbol{I}_3 & \boldsymbol{0}_3 \\ |{}^{o}\boldsymbol{p}_r|_{\times} & \boldsymbol{I}_3 \end{bmatrix}$ 

\hspace{1.0cm}${}^r\boldsymbol{F}_o = \begin{bmatrix} {}^r\boldsymbol{f}_o \\ {}^r\boldsymbol{t}_o \end{bmatrix} = \begin{bmatrix} \boldsymbol{f}_r + \boldsymbol{f}_h \\ \boldsymbol{t}_r + {}^r\boldsymbol{p}_h \times \boldsymbol{f}_h + \boldsymbol{t}_h\end{bmatrix}$     

$\\$
The variable ${}^{r}\boldsymbol{F}_o$ denotes the wrench exerted on the object with respect to the robot frame and can be directly estimated based on the robot's sensory feedback and measurement capabilities.

By substituting (\ref{eq_5}) and (\ref{eq_6}) into (\ref{eq_1}) and manipulating the result with the parallel axis theorem, the object dynamics (\ref{eq_1}) can be represented with respect to the robot frame $\{r\}$ as follows:
\begin{eqnarray}
	\label{eq_7}
    \begin{array}{l}     
     \begin{bmatrix} \boldsymbol{m}_o \{ \dot{\boldsymbol{v}}_r + \dot{ \boldsymbol{\omega}}_o \times {}^{r}\boldsymbol{p}_o +  \boldsymbol{\omega}_o \times ( \boldsymbol{\omega}_o \times {}^{r}\boldsymbol{p}_o ) \} -\boldsymbol{m}_o \boldsymbol{g} \\
                     {}^{r}\boldsymbol{J}_o \dot{ \boldsymbol{\omega}}_o +  \boldsymbol{\omega}_o \times {}^{r}\boldsymbol{J}_o  \boldsymbol{\omega}_o + m_o {}^{r}\boldsymbol{p}_o \times (\dot{\boldsymbol{v}}_r - \boldsymbol{g}) \end{bmatrix}      \\
     = \begin{bmatrix} {}^{r}\boldsymbol{f}_o \\ {}^{r}\boldsymbol{t}_o  \end{bmatrix}

     \end{array}
\end{eqnarray}
where ${}^r\boldsymbol{J}_o$ is the object inertia matrix expressed in the robot frame.

\subsection{Unknown Object Parameter Estimation}
For human-robot collaboration tasks, obtaining human motion and wrench measurements requires that the human participant be equipped with additional sensors for motion tracking or force/torque feedback \cite{Cehajic}.
However, in most real-world scenarios, it is challenging to use additional sensors, leading to limited available human information. Therefore, we estimated the object parameters in a scenario without human motion information and wrench measurements.
In this study, the objective of the parameter estimation process is not to precisely obtain the real parameters of the object. Instead, the aim is to accurately recognize the intended human motion by compensating for the effects of the object dynamics, thereby enabling safe and precise human-robot collaboration.
Consequently, in the proposed approach, we focus on estimating the effective object parameters that the robot needs during the collaborative task.

According to (\ref{eq_2}) and (\ref{eq_6}), the wrench $^rF_o$ can be represented as follows:
\begin{eqnarray}
	\label{eq_8}
    \begin{array}{l} 
     {}^{r}\boldsymbol{F}_o = 
     \begin{bmatrix} \boldsymbol{I}_6 & {}^r\boldsymbol{G}_h \end{bmatrix}         
     \begin{bmatrix} \boldsymbol{F}_r \\ \boldsymbol{F}_h  \end{bmatrix} 
     = \boldsymbol{F}_r + {}^r\boldsymbol{G}_h \boldsymbol{F}_h \\
     \end{array}
\end{eqnarray}
where

\hspace{1.0cm}${}^r\boldsymbol{G}_h = \begin{bmatrix} \boldsymbol{I}_3 & \boldsymbol{0}_3 \\ |{}^{r}\boldsymbol{p}_h|_{\times} & \boldsymbol{I}_3 \end{bmatrix}$     

$\\$
As previously mentioned, since we do not account for the human wrench during the estimation process, $F_h$ is set to zero as follows:
\begin{eqnarray}
	\label{eq_9}
    \begin{array}{l}     
     {}^{r}\boldsymbol{F}_o = \boldsymbol{F}_r    
    \end{array}
\end{eqnarray}

By substituting (\ref{eq_9}) into (\ref{eq_7}), the object dynamics affecting the robot can be represented as follows:
\begin{eqnarray}
	\label{eq_10}
    \begin{array}{l}     
    \boldsymbol{F}_r = \boldsymbol{A}(\boldsymbol{a},\dot{ \boldsymbol{\omega}}, \boldsymbol{\omega},\boldsymbol{g}) \boldsymbol{\phi}_{eff}    
    \end{array}
\end{eqnarray}
where $\boldsymbol{A}(\boldsymbol{a},\dot{ \boldsymbol{\omega}}, \boldsymbol{\omega}, \boldsymbol{g})$ is a data matrix containing the linear and angular accelerations and angular velocities \cite{Kurdas}. The parameter vector $\boldsymbol{\theta} \in \mathbb{R}^{n_{\theta}}, n_{\theta}=10$ consists of the object's dynamic parameters and is defined as follows:
\begin{eqnarray}
	\label{eq_11}
    \begin{array}{l}     
    \boldsymbol{\phi} = \begin{bmatrix} \boldsymbol{m}_o & \boldsymbol{m}_o {^{r}\boldsymbol{p}_o}^T & {}^{r}\boldsymbol{J}_{o,vec} \end{bmatrix}^T 
    \end{array}
\end{eqnarray}
where $\boldsymbol{m}_o$ is the object mass, ${}^r\boldsymbol{p}_o$ is the center of mass between the robot and object frames, and $^{r}J_{o,vec}$ is the moment of inertia, with $^{r}J_{o,vec} = \begin{bmatrix} {}^{r}\boldsymbol{J}_{o,xx} & {}^{r}\boldsymbol{J}_{o,xy} & {}^{r}\boldsymbol{J}_{o,xz} & {}^{r}\boldsymbol{J}_{o,yy} & {}^{r}\boldsymbol{J}_{o,yz} & {}^{r}\boldsymbol{J}_{o,zz} \end{bmatrix}^T$.

The extended Kalman filter (EKF) is adopted as an estimator because $\boldsymbol{A}(\boldsymbol{a},\dot{ \boldsymbol{\omega}}, \boldsymbol{\omega},\boldsymbol{g})$ is time varying. With the filter state $\hat{\boldsymbol{x}}(k) = \hat{\boldsymbol{\phi}}_{eff}$, the prediction of the EKF is formulated as follows:
\begin{eqnarray}
	\label{eq_12}
    \begin{array}{l}     
    \hat{\boldsymbol{\phi}}_{eff}(k|k-1) = \hat{\boldsymbol{\phi}}_{eff}(k-1)
    \end{array}
\end{eqnarray}
\begin{eqnarray}
	\label{eq_13}
    \begin{array}{l}     
    \boldsymbol{P}(k|k-1) = \boldsymbol{P}(k-1) + \boldsymbol{Q}(k)
    \end{array}
\end{eqnarray}
where $\boldsymbol{Q} \in \mathbb{R}^{n_{\boldsymbol{\theta}} \times n_{\theta}}$ is the process noise covariance matrix. Then, the correction process is formulated with the effective object dynamics as follows:
\begin{eqnarray}
	\label{eq_14}
    \begin{array}{l}     
    \boldsymbol{K}(k) = \boldsymbol{P}(k|k-1)\boldsymbol{H}^T(k) \times \\ 
    \begin{bmatrix} \boldsymbol{H}(k)\boldsymbol{P}(k|k-1)\boldsymbol{H}^T(k) + \boldsymbol{R}(k)\end{bmatrix}^{-1}
    \end{array}
\end{eqnarray}
\begin{eqnarray}
	\label{eq_15}
    \begin{array}{l}     
    \hat{\boldsymbol{\phi}}_{eff}(k) = \hat{\boldsymbol{\phi}}_{eff}(k|k-1) \\
    + \boldsymbol{K}(k) \{ \bar{\boldsymbol{F}}_r(k) - \boldsymbol{A}(\boldsymbol{a},\dot{ \boldsymbol{\omega}}, \boldsymbol{\omega},\boldsymbol{g}) \hat{\boldsymbol{\phi}}_{eff}(k|k-1) \}
    \end{array}
\end{eqnarray}
\begin{eqnarray}
	\label{eq_16}
    \begin{array}{l}     
    \boldsymbol{P}(k) = (\boldsymbol{I}-\boldsymbol{K}(k)\boldsymbol{H}(k))\boldsymbol{P}(k|k-1)
    \end{array}
\end{eqnarray}
where $\boldsymbol{R} \in \mathbb{R}^{6 \times 6}$ is the measurement noise covariance matrix, $\boldsymbol{P}$ is the estimated error covariance matrix, and $\boldsymbol{H}(k) = \frac{\partial}{\partial \boldsymbol{x}}\boldsymbol{A}(\boldsymbol{a},\dot{ \boldsymbol{\omega}}, \boldsymbol{\omega},\boldsymbol{g}) \hat{\boldsymbol{\phi}}_{eff}|_{\hat{\boldsymbol{x}}(k|k-1)}$ is obtained by backward differentiation. $\bar{\boldsymbol{F}}_r(k)$ is directly measured based on the robot sensory feedback, and $\hat{\boldsymbol{\phi}}_{eff}$ is the value estimated according to this filter.

With the estimated effective object dynamics, the real object dynamics can be expressed according to (\ref{eq_8}) as follows:
\begin{eqnarray}
	\label{eq_17}
    \begin{array}{l}     
    \boldsymbol{A}(\boldsymbol{a},\dot{ \boldsymbol{\omega}}, \boldsymbol{\omega},\boldsymbol{g}) \boldsymbol{\phi}_{eff} + {}^{r}\boldsymbol{G}_h \boldsymbol{F}_h = \boldsymbol{A}(\boldsymbol{a},\dot{ \boldsymbol{\omega}}, \boldsymbol{\omega},\boldsymbol{g}) \boldsymbol{\phi}_{real}
    \end{array}
\end{eqnarray}

Here, $\boldsymbol{F}_h$ denotes the intended human motion transmitted to the robot through ${}^r\boldsymbol{G}_h$. Therefore, the real object dynamics are expressed as a sum of the effective object dynamics and the intended human motion projected by the object geometry.

\section{Object-Aware Impedance Controller}
An object-aware impedance controller is designed based on the estimated effective object dynamics for human-robot collaborative mobile manipulation tasks.

\subsection{Dynamics of the Mobile Manipulator}
The dynamics of the nonholonomic mobile manipulator are briefly reviewed \cite{Kim}.
The mobile manipulator consists of a four-wheel differential-drive mobile base and manipulator, as depicted in Fig. \ref{Fig_1}.
The generalized coordinates include 3-degree-of-freedom (DOF) rigid body motion ($ \boldsymbol{\zeta}$) due to the presence of a floating base, wheel angular motion ($\boldsymbol{q}_b$) and arm joint motion ($\boldsymbol{q}_m$) and are written as follows:
\begin{eqnarray}
	\label{eq_18}
    \begin{array}{l}     
     \boldsymbol{q}= \begin{bmatrix}  \boldsymbol{\zeta}^T \quad \boldsymbol{q}^{T}_b \quad \boldsymbol{q}^{T}_m \end{bmatrix}^T
    \end{array}
\end{eqnarray}
where $ \boldsymbol{\zeta} = \begin{bmatrix} {x}_c \quad {y}_c \quad {\phi} \end{bmatrix}^T \in \mathbb{R}^3$, $\boldsymbol{q}_b = \begin{bmatrix} \theta_r \quad \theta_l \end{bmatrix}^T \in \mathbb{R}^2$ and $\boldsymbol{q}_m \in \mathbb{R}^{n_m}$. Note that $n_b$ has only two components despite the presence of the four wheels because the left and right pairs of wheels are dependent on each other. As the manipulator used in our system has 7-DOF motion, the total DOF of the generalized coordinates is $n=n_{ \boldsymbol{\zeta}} + n_b + n_m = 3 + 2 + 7 = 12$.

The differential-drive mobile base is characterized by two nonholonomic constraints: no lateral slip and pure rolling motion. The constraint matrix $\boldsymbol{A}_c(\boldsymbol{q})$ can be represented as follows:
\begin{eqnarray}
	\label{eq_19}
    \begin{array}{l}     
     \boldsymbol{A}_c(\boldsymbol{q})\dot{\boldsymbol{q}} = 0
    \end{array}
\end{eqnarray}
where $\boldsymbol{A}_c(\boldsymbol{q}) \in \mathbb{R}^{n_{ \boldsymbol{\zeta}} \times n}$.
The transformation matrix $\boldsymbol{S}(\boldsymbol{q})$ is in the null space of the constraint matrix $\boldsymbol{A}_c(\boldsymbol{q})$ \cite{Dhaouadi}.
\begin{eqnarray}
	\label{eq_20}
    \begin{array}{l}     
     \dot{\boldsymbol{q}} = \boldsymbol{S}(\boldsymbol{q}) \boldsymbol{\eta}
    \end{array}
\end{eqnarray}
where $\boldsymbol{S}(\boldsymbol{q}) \in \mathbb{R}^{n \times (n_b \times n_m)}$ satisfies $\boldsymbol{A}_c(\boldsymbol{q})\boldsymbol{S}(\boldsymbol{q})=0$ and $\boldsymbol{\eta} = \begin{bmatrix} \dot{q}^{T}_b \quad \dot{q}^{T}_m \end{bmatrix}^T \in \mathbb{R}^{n_b+n_m}$ is the reduced velocity vector of the generalized coordinate of the robot. The Jacobian matrix between the Cartesian velocity space and the reduced joint velocity space can be derived as follows:
\begin{eqnarray}
	\label{eq_21}
    \begin{array}{l}     
     \boldsymbol{J}(\boldsymbol{q}) = \tilde{\boldsymbol{J}}(\boldsymbol{q})\boldsymbol{S}(\boldsymbol{q})
    \end{array}
\end{eqnarray}
where $\boldsymbol{J} \in \mathbb{R}^{6 \times (n_b + n_m)}$.
The nonholonomic constraint can be included in the equation of motion of the mobile manipulator by introducing a Lagrangian multiplier as follows:
\begin{eqnarray}
	\label{eq_22}
    \begin{array}{l}     
     \tilde{\boldsymbol{M}}(\boldsymbol{q}) \ddot{\boldsymbol{q}} + \tilde{\boldsymbol{C}}(\boldsymbol{q},\dot{\boldsymbol{q}})\dot{\boldsymbol{q}} + \tilde{\boldsymbol{g}}(\boldsymbol{q}) = \tilde{ \boldsymbol{\tau}}_{input} + \tilde{ \boldsymbol{\tau}}_{ext} + \boldsymbol{A}^{T}_c(\boldsymbol{q})\boldsymbol{\lambda}
    \end{array}
\end{eqnarray}
where $\tilde{ \boldsymbol{\tau}}_{input} \in \mathbb{R}^n$ and $\tilde{ \boldsymbol{\tau}}_{ext} \in \mathbb{R}^n$ are the joint torque input vector and the external disturbance vector, respectively. $\tilde{\boldsymbol{M}}(q) \in \mathbb{R}^{n \times n}$, $\tilde{\boldsymbol{C}}(\boldsymbol{q},\dot{\boldsymbol{q}}) \in \mathbb{R}^{n \times n}$, and $\tilde{\boldsymbol{g}}(\boldsymbol{q}) \in \mathbb{R}^{n}$ are the inertia matrix, Coriolis and centrifugal matrix, and gravity vector, respectively. $\boldsymbol{\lambda} \in \mathbb{R}^{n_{ \boldsymbol{\zeta}}}$ is the Lagrangian multiplier for the nonholonomic constraints.

By substituting (\ref{eq_20}) and its time derivative into (\ref{eq_22}) and multiplying by $\boldsymbol{S}^T (\boldsymbol{q})$ on both sides of the equation, the nonconstrained equation of motion can be reformulated as follows:
\begin{eqnarray}
	\label{eq_23}
    \begin{array}{l}     
     \boldsymbol{M}(\boldsymbol{q}) \dot{\boldsymbol{\eta}} + \boldsymbol{C}(q,\dot{q})\boldsymbol{\eta} + \boldsymbol{g}(\boldsymbol{q}) =  \boldsymbol{\tau}_{input} +  \boldsymbol{\tau}_{ext} 
    \end{array}
\end{eqnarray}

where

\hspace{0.5cm} $\boldsymbol{M}(\boldsymbol{q}) = \boldsymbol{S}(\boldsymbol{q})^T \tilde{\boldsymbol{M}}(\boldsymbol{q})\boldsymbol{S}(\boldsymbol{q})$

\hspace{0.5cm} $\boldsymbol{C}(\boldsymbol{q},\dot{\boldsymbol{q}}) = \boldsymbol{S}(\boldsymbol{q})^T \begin{bmatrix} \tilde{\boldsymbol{M}}(\boldsymbol{q}) \dot{\boldsymbol{S}}(\boldsymbol{q}) + \tilde{\boldsymbol{C}}(q,\dot{q})S(q) \end{bmatrix}$

\hspace{0.5cm} $\boldsymbol{g}(\boldsymbol{q}) = \boldsymbol{S}(\boldsymbol{q})^T \tilde{\boldsymbol{g}}(\boldsymbol{q})$

\hspace{0.5cm} $ \boldsymbol{\tau}_{input} = \boldsymbol{S}(\boldsymbol{q})^T \tilde{ \boldsymbol{\tau}}$

\hspace{0.5cm} $ \boldsymbol{\tau}_{ext} = \boldsymbol{S}(\boldsymbol{q})^T \tilde{ \boldsymbol{\tau}}_{ext}$ 

$\\$
Consequently, the 3-DOF rigid body motion is eliminated from the state vector of the reformulated equation, enabling the use of the nonholonomic constraints.

\subsection{QP-Based Whole-Body Controller}
Due to the redundancy of the whole-body robot, quadratic programming (QP) optimization can be introduced to develop a multipriority control framework. The redundancy is due to the difference between the number of joints and the number of Cartesian DOFs, and the Jacobian can be introduced to address this redundancy.
The QP optimization can be formulated as follows:
\begin{eqnarray}
	\label{eq_24}
    \begin{array}{l}     
     \displaystyle \min_{\dot{\boldsymbol{\eta}}} \quad \boldsymbol{w}_{j} \lVert \dot{\boldsymbol{\eta}} - \dot{\boldsymbol{\eta}}_{d1} \rVert _2 + 
                    \boldsymbol{w}_{t} \lVert \boldsymbol{\ddot{x}}_{d1} - J_1 \dot{\boldsymbol{\eta}} - \dot{J}_1 \boldsymbol{\eta} \rVert _2 \\     
     s.t. \quad \dot{\boldsymbol{\eta}} = \dot{\boldsymbol{\eta}}_{d0} \\
     \quad \quad \, \, \boldsymbol{\ddot{x}}_{d0} = J_0 \dot{\boldsymbol{\eta}} + \dot{J}_0 \boldsymbol{\eta} \\
     \quad \quad \, \, \underbar{$\boldsymbol{\eta}$}_{l} < \dot{\boldsymbol{\eta}} < \bar{\boldsymbol{\eta}}_{u}
    \end{array}
\end{eqnarray}
where $\boldsymbol{w_j}$ and $\boldsymbol{w_t}$ are the weight vectors in the cost function. $\dot{\boldsymbol{\eta}}_{d1}$ and $\boldsymbol{\ddot{x}}_{d1}$ are the joint and Cartesian space desired accelerations introduced in the cost function, and $\dot{\boldsymbol{\eta}}_{d0}$ and $\boldsymbol{\ddot{x}}_{d0}$ are the desired accelerations introduced in the constraints of the QP formulation. $\underbar{$\boldsymbol{\eta}$}_{l}$ and $\bar{\boldsymbol{\eta}}_{u}$ are the lower and upper limits of the joint acceleration, respectively.

If the solution of~(\ref{eq_24}) exits, it should satisfy the equality and inequality constraints.
On the other hand, the goal of the optimization problem is to find a solution that minimizes the resultant cost of the loss function under the given constraints. Consequently, it is not necessary for each term in the loss function to converge to zero.
With this perspective, the tasks designated by the constraints must be satisfied, and they have the highest priority (priority = 0), while the tasks in the loss function have lower priority (priority = 1).
In summary, the QP optimization process can be formulated by selecting tasks based on the candidates, such as priority 0 tasks $(\dot{\boldsymbol{\eta}}_{d0}, \boldsymbol{\ddot{x}}_{d0}), (\boldsymbol{\eta}_l, \boldsymbol{\eta}_u)$ and/or priority 1 tasks ($\dot{\boldsymbol{\eta}}_{d1}, \boldsymbol{\ddot{x}}_{d1}$).

In the joint space, the desired acceleration can be obtained through a PD controller as follows:
\begin{eqnarray}
	\label{eq_25}
    \begin{array}{l}     
     \dot{\boldsymbol{\eta}}_{di} =\begin{bmatrix} \boldsymbol{\ddot{q}}_b \\ \boldsymbol{\ddot{q}}_m \end{bmatrix}
                    -\boldsymbol{K}_{P\boldsymbol{\eta}}(\begin{bmatrix} \boldsymbol{q}_b \\ \boldsymbol{q}_m \end{bmatrix} - \begin{bmatrix} \boldsymbol{q}_b \\ \boldsymbol{q}_m \end{bmatrix}_d)  \\
                    -\boldsymbol{K}_{D\boldsymbol{\eta}}(\begin{bmatrix} \dot{\boldsymbol{q}}_b \\ \dot{\boldsymbol{q}}_m \end{bmatrix} - \begin{bmatrix} \dot{\boldsymbol{q}}_b \\ \dot{\boldsymbol{q}}_m \end{bmatrix}_d) 
    \end{array}
\end{eqnarray}
where $i=0, 1$ denotes the priorities. $\boldsymbol{K}_{P\boldsymbol{\eta}}$ and $\boldsymbol{K}_{D\boldsymbol{\eta}}$ are the proportional and derivative gains in joint space, respectively, and $\begin{bmatrix} \boldsymbol{q}^T_b & \boldsymbol{q}^T_m \end{bmatrix}^T_d$ denotes the desired joint angle.
Additionally, the desired acceleration in Cartesian space can be obtained as follows:
\begin{eqnarray}
	\label{eq_26}
    \begin{array}{l}     
     \ddot{\boldsymbol{x}}_{d1} = \ddot{\boldsymbol{x}} -\boldsymbol{K}_{P\boldsymbol{x}}(\boldsymbol{x}-\boldsymbol{x}_d) -\boldsymbol{K}_{D\boldsymbol{x}}(\dot{\boldsymbol{x}} - \dot{\boldsymbol{x}}_d)          
    \end{array}
\end{eqnarray}
where $\boldsymbol{K}_{P\boldsymbol{x}}$ and $\boldsymbol{K}_{D\boldsymbol{x}}$ are the proportional and derivative gains in Cartesian space, respectively, and $\boldsymbol{x}_d, \dot{\boldsymbol{x}}_d$ are the desired pose and velocity, respectively. In this paper, instead of (\ref{eq_26}), we adopt a Cartesian space object-aware impedance controller.

According to the solution shown in (\ref{eq_24}), $\dot{\boldsymbol{\eta}}^*$, the desired input torque for the real robot, can be obtained with the feedback linearization technique as follows:
\begin{eqnarray}
	\label{eq_27}
    \begin{array}{l}     
      \boldsymbol{\tau}_{input} = \boldsymbol{M}(\boldsymbol{q}) \dot{\boldsymbol{\eta}}^* + \boldsymbol{C}(\boldsymbol{q},\dot{\boldsymbol{q}})\boldsymbol{\eta} + \boldsymbol{g}(\boldsymbol{q}) -  \boldsymbol{\tau}_{ext} 
    \end{array}
\end{eqnarray}

\begin{figure}[t]\centering
    \centering
    \includegraphics[scale=0.6]{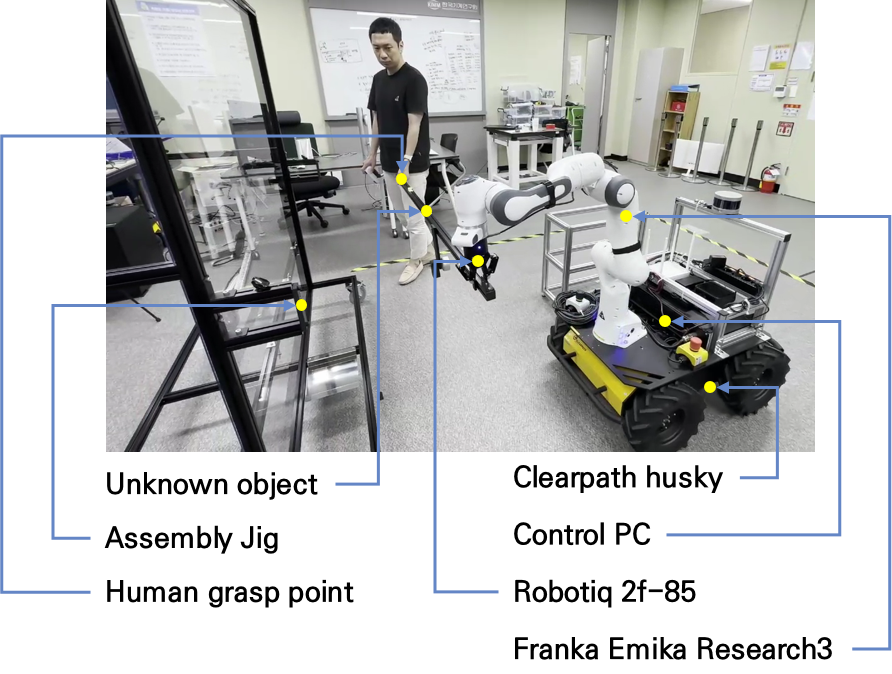}    
    \caption{Mobile manipulator system and assembly jig for the human-robot collaborative task scenario.}
    \label{Fig_2}
\end{figure}

\subsection{Object-Aware Impedance Controller Formulation}
For a human-robot collaborative task with an object, the external force is determined by the real object dynamics as follows:
\begin{eqnarray}
	\label{eq_28}
    \begin{array}{l}     
      \boldsymbol{\tau}_{ext} = \boldsymbol{J}^{T}\boldsymbol{F}_{ext} = \boldsymbol{J}^{T}\boldsymbol{A}(\boldsymbol{a},\dot{ \boldsymbol{\omega}}, \boldsymbol{\omega},\boldsymbol{g}) \boldsymbol{\phi}_{real}
    \end{array}
\end{eqnarray}

By substituting (\ref{eq_17}) into (\ref{eq_23}), the equation of motion considering the object dynamics can be represented as follows:
\begin{eqnarray}
	\label{eq_29}
    \begin{array}{l}     
     \boldsymbol{M}(\boldsymbol{q}) \dot{\boldsymbol{\eta}} + \boldsymbol{C}(\boldsymbol{q},\dot{\boldsymbol{q}})\boldsymbol{\eta} + \boldsymbol{g}(\boldsymbol{q}) =  \\
     \quad \quad \quad \quad  \boldsymbol{\tau}_{input} + \boldsymbol{J}^{T}\{ {}^{r}\boldsymbol{G}_h \boldsymbol{F}_h + \boldsymbol{A}(\boldsymbol{a},\dot{ \boldsymbol{\omega}}, \boldsymbol{\omega},\boldsymbol{g}) \boldsymbol{\phi}_{eff} \}
    \end{array}
\end{eqnarray}
The external forces acting on the robot include the human wrench, which is exerted through the object, and the effective object dynamics that the robot endures under the specific configuration.
If there is no object, $^r\boldsymbol{G}_h$ becomes an identity matrix, reducing the external forces to $\boldsymbol{F}_{ext}=^{r}\boldsymbol{G}_h\boldsymbol{F}_h=\boldsymbol{F}_h$, signifying that the intended human motion is applied directly to the robot. On the other hand, in the absence of human-robot collaboration, in which only the robot performs the task, the robot bears all the object dynamics. Hence,  (\ref{eq_17}) becomes $\boldsymbol{F}_{ext} = \boldsymbol{A}(\boldsymbol{a},\dot{ \boldsymbol{\omega}}, \boldsymbol{\omega},\boldsymbol{g}) \hat{\boldsymbol{\phi}}_{eff} =  \boldsymbol{A}(\boldsymbol{a},\dot{ \boldsymbol{\omega}}, \boldsymbol{\omega},\boldsymbol{g}) \hat{\boldsymbol{\phi}}_{real}$, representing that the real object dynamics are manifested as external forces.

For human-robot collaborative tasks, the target impedance can be designed as follows:
\begin{eqnarray}
	\label{eq_30}
    \begin{array}{l}     
    \boldsymbol{M}_e (\ddot{\boldsymbol{x}} - \ddot{\boldsymbol{x}}_{d} ) + \boldsymbol{C}_e (\dot{\boldsymbol{x}} - \dot{\boldsymbol{x}}_{d} ) + \boldsymbol{K}_e (\boldsymbol{x}-\boldsymbol{x}_{d}) = \boldsymbol{F}_h
    \end{array}
\end{eqnarray}

The crucial point is that the external force in the desired impedance is $\boldsymbol{F}_h$ and not $^{r}\boldsymbol{G}_h \boldsymbol{F}_h$. When the human and the robot are cooperatively holding an object, the desired impedance represents the relationship between the robot motion and the intended human motion, excluding the influence of the object.
Note that if the stiffness of the desired impedance is set to zero, the robot can actively follow the human operator based on the estimated human motion.

To implement the impedance controller using the QP formulation, the Cartesian space task $\ddot{\boldsymbol{x}}_{d0}$ with priority 0 can be expressed as follows:
\begin{eqnarray}
	\label{eq_31}
    \begin{array}{l}     
      \ddot{\boldsymbol{x}}_{d0} = \ddot{\boldsymbol{x}}_{d} - \boldsymbol{M}^{-1}_e\boldsymbol{C}_e (\dot{\boldsymbol{x}} - \dot{\boldsymbol{x}}_{d} ) - \boldsymbol{M}^{-1}_e\boldsymbol{K}_e (\boldsymbol{x}-\boldsymbol{x}_{d}) \\ + \boldsymbol{M}^{-1}_e \boldsymbol{F}_h
    \end{array}
\end{eqnarray}
The above equation is similar to (\ref{eq_26}); however, instead of a PD controller, an impedance controller including the desired inertia and external force is employed.

By assigning the impedance controller as the highest priority task and solving the QP problem, the object-aware impedance controller can be represented as follows:
\begin{eqnarray}
	\label{eq_32}
    \begin{array}{l}     
    \dot{\boldsymbol{\eta}} = \boldsymbol{J}^{+}_0 \{ \ddot{\boldsymbol{x}}_{d} + \boldsymbol{M}^{-1}_e \boldsymbol{F}_h \\
    \quad \, \, \, -\boldsymbol{M}^{-1}_e (\boldsymbol{C}_e(\dot{\boldsymbol{x}}-\dot{\boldsymbol{x}}_{d}) + \boldsymbol{K}_e (\boldsymbol{x} - \boldsymbol{x}_{d})) - \dot{\boldsymbol{J}}_0\dot{\boldsymbol{\eta}}  \}
    \end{array}
\end{eqnarray}
where $\boldsymbol{J}^{+}_0$ is the pseudoinverse of $\boldsymbol{J}_0$. Without loss of generality, the existence of a feasible solution for the impedance controller can be guaranteed due to the its highest priority.
Therefore, the desired input torque can be represented by substituting (\ref{eq_32}) into (\ref{eq_29}) as follows:
\begin{eqnarray}
	\label{eq_33}
    \begin{array}{l}     
     \boldsymbol{\tau}_{input} = \\
        \quad \boldsymbol{M}\boldsymbol{J}^{+}_0 \{ \ddot{\boldsymbol{x}}_{d} + \boldsymbol{M}^{-1}_e \boldsymbol{F}_h \\
        \quad + \boldsymbol{M}^{-1}_e (-\boldsymbol{C}_e(\dot{\boldsymbol{x}}-\dot{\boldsymbol{x}}_{d}) - \boldsymbol{K}_e (\boldsymbol{x} - \boldsymbol{x}_{d})) -\dot{\boldsymbol{J}}_0\dot{\boldsymbol{q}} \} \\
        \quad +\boldsymbol{C}(\boldsymbol{q},\dot{\boldsymbol{q}})\boldsymbol{\eta} + \boldsymbol{g}(\boldsymbol{q}) \\
        \quad -\boldsymbol{J}^T \{^{r}\boldsymbol{G}_h \boldsymbol{F}_h + \boldsymbol{A}(\boldsymbol{a},\dot{ \boldsymbol{\omega}}, \boldsymbol{\omega},\boldsymbol{g}) \hat{\boldsymbol{\phi}}_{eff}\}
    \end{array}
\end{eqnarray}

Alternatively, by gathering the terms related to $F_h$, (\ref{eq_33}) can be represented as follows:
\begin{eqnarray}
	\label{eq_34}
    \begin{array}{l}     
     \boldsymbol{\tau}_{input} = \\
        \quad \boldsymbol{M}\boldsymbol{J}^{+}_0 \{ \ddot{\boldsymbol{x}}_{d}  \\
        \quad + \boldsymbol{M}^{-1}_e (-\boldsymbol{C}_e(\dot{\boldsymbol{x}}-\dot{\boldsymbol{x}}_{d}) - \boldsymbol{K}_e (\boldsymbol{x} - \boldsymbol{x}_{d})) -\dot{\boldsymbol{J}}_0\dot{\boldsymbol{q}} \} \\
        \quad +\boldsymbol{J}^T_0 \boldsymbol{M}_F \boldsymbol{F}_h \\
        \quad +\boldsymbol{C}(\boldsymbol{q},\dot{\boldsymbol{q}})\boldsymbol{\eta} + \boldsymbol{g}(\boldsymbol{q}) -\boldsymbol{J}^T \boldsymbol{A}(\boldsymbol{a},\dot{ \boldsymbol{\omega}}, \boldsymbol{\omega},\boldsymbol{g}) \hat{\boldsymbol{\phi}}_{eff}             
    \end{array}
\end{eqnarray}
where $\boldsymbol{M}_F = (\boldsymbol{M}_x \boldsymbol{M}^{-1}_e - {^{r}\boldsymbol{G}_h})$ and $\boldsymbol{M}_x = \boldsymbol{J}^{+T}_0 \boldsymbol{M} \boldsymbol{J}^{+}_0$, which is the inertia in Cartesian space \cite{Bussmann}.
For a general impedance controller, external force feedback can be avoided when inertia shaping is not performed, as expressed by $\boldsymbol{M}_e = \boldsymbol{M}_{\boldsymbol{x}}$ \cite{Ott}. 
However, in scenarios in which an object is present between the robot and human, avoiding the external forces feedback requires specific inertia shaping with considering object geometry, which is expressed as $\boldsymbol{M}_e = {}^{r}\boldsymbol{G}^{-1}_h \boldsymbol{M}_x$, thereby resulting in $\boldsymbol{M}_F = 0$. 
Accordingly, the desired impedance is formulated as follows:
\begin{eqnarray}
	\label{eq_35}
    \begin{array}{l}     
    ^{h}\boldsymbol{G}_r \boldsymbol{M}_x (\ddot{\boldsymbol{x}} - \ddot{\boldsymbol{x}}_{d} ) + \boldsymbol{C}_e (\dot{\boldsymbol{x}} - \dot{\boldsymbol{x}}_{d} ) + \boldsymbol{K}_e (\boldsymbol{x}-\boldsymbol{x}_{d}) \\ = \boldsymbol{F}_h
    \end{array}
\end{eqnarray}
where $^{r}\boldsymbol{G}^{-1}_h = {}^h\boldsymbol{G}_r$. 
However, understanding the intended human motion, denoted as $\boldsymbol{F}_h$, necessitates the use of external force measurements. Consequently, ifurther shaping of the desired inertia is employed to enable efficient interactions between the human and the robot through the object as follows:
\begin{eqnarray}
	\label{eq_36}
    \begin{array}{l}     
    \boldsymbol{M}_e = (^r\boldsymbol{G}_h + \boldsymbol{P}_d)^{-1} \boldsymbol{M}_x 
    \longrightarrow  \boldsymbol{M}_F = \boldsymbol{P}_d
    \end{array}
\end{eqnarray}
where $\boldsymbol{P}_d \in \mathbb{R}^{6 \times 6}$ is the desired weighting matrix of the external wrenches, which is determined in the subsequent section.

\begin{figure*}[t]\centering
    \centering
    \includegraphics[scale=0.6]{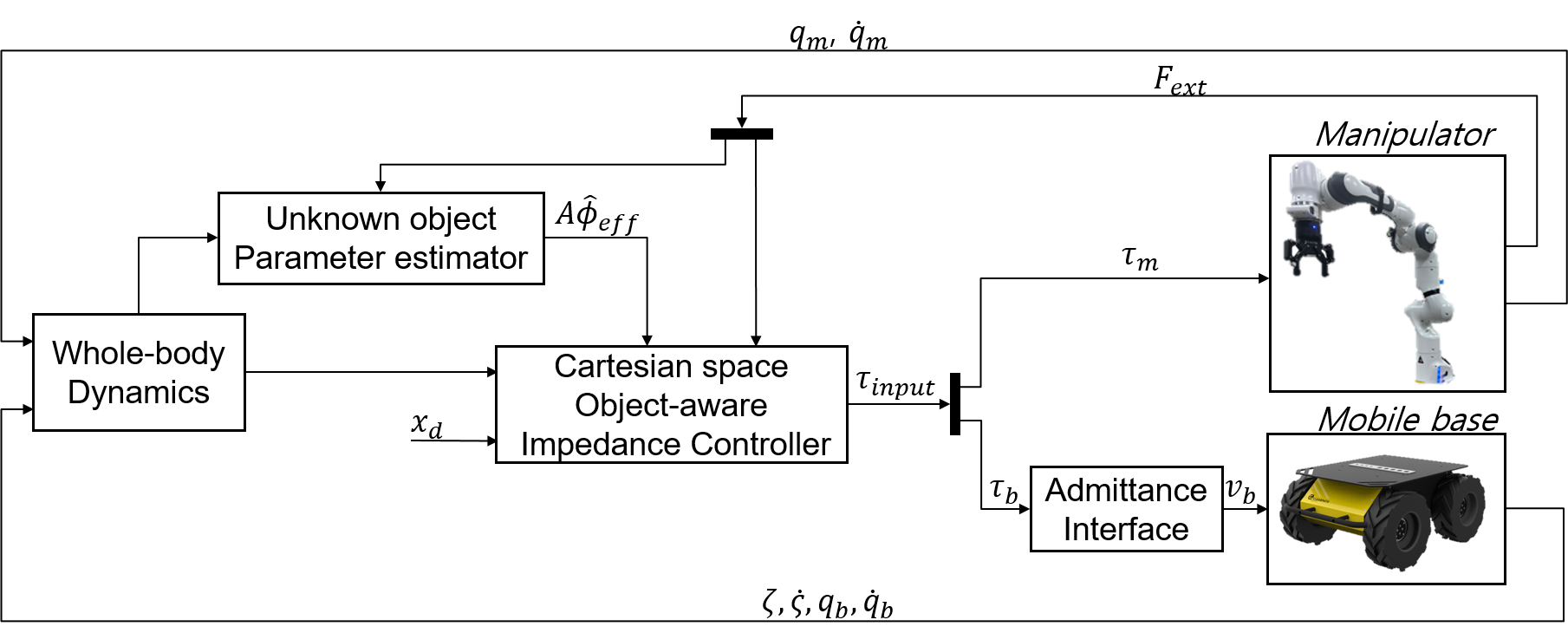}    
    \caption{Schematic diagram of the unknown object parameter estimation process and object-aware impedance controller.}
    \label{Fig_3}
\end{figure*}

\section{Experiments}
The mobile manipulator was realized, as illustrated in Fig.~\ref{Fig_2}, to experimentally verify the performance of the proposed method in human-robot collaborative task scenarios by integrating a 7-DOF torque-controlled manipulator, Research3, manufactured by Fanka Emika, and a velocity-controlled four-wheel mobile base, Husky, manufactured by Clearpath Robotics. The mobile base is operated by a differential drive maneuver, and the left and right pairs of wheels are dependent on each other, resulting in 2-DOF motion. A 1-DOF two-fingered gripper, 2F-85 by Robotiq, is attached to the end-effector of the robot arm.
The control algorithms are implemented on an Intel Core i7 4.7 GHz with 16 GB RAM with a robot operating system (ROS) interface and PREEMPT\_RT kernel.

A schematic diagram of the proposed human-robot collaborative task, including the unknown object parameter estimation process and the Cartesian space object-aware impedance controller, is shown in Fig.~\ref{Fig_3}. The parameter estimation process is carried out during the estimation phase, and after the parameters are estimated, the object dynamics, $\boldsymbol{A}\hat{\boldsymbol{\phi}}_{eff}$, can be transmitted to the controller. To enhance the estimation precision, a perturbation signal is introduced into the null space of $x_d$, as detailed in the next section.
The measurements are the state of the manipulator, $\boldsymbol{q}$, $\dot{\boldsymbol{q}}$, and mobile base, $\boldsymbol{q}_b$, $\dot{\boldsymbol{q}}_b$, and the position and velocity of the mobile manipulator, $ \boldsymbol{\zeta}$, ${\dot{ \boldsymbol{\zeta}}}$, are provided by localization algorithm \cite{Husky}.
The external force, $\boldsymbol{F}_{ext}$, exerted on the end effector is also estimated based on the torque sensors equipped at each joint of the manipulator.
The desired torque input computed by our controller is directly transmitted to the robot arm, while for the velocity-controlled mobile base, the torque inputs are transformed into velocity inputs with the admittance interface ~\cite{Dietrich}. The manipulator is controlled in real time with a 1 kHz control frequency, and the mobile base is controlled by ROS topic communication with a 10 Hz sampling rate.

\subsection{Unknown Object Parameter Estimation Results}
\begin{figure*}[t]\centering
    \centering
    \includegraphics[scale=0.55]{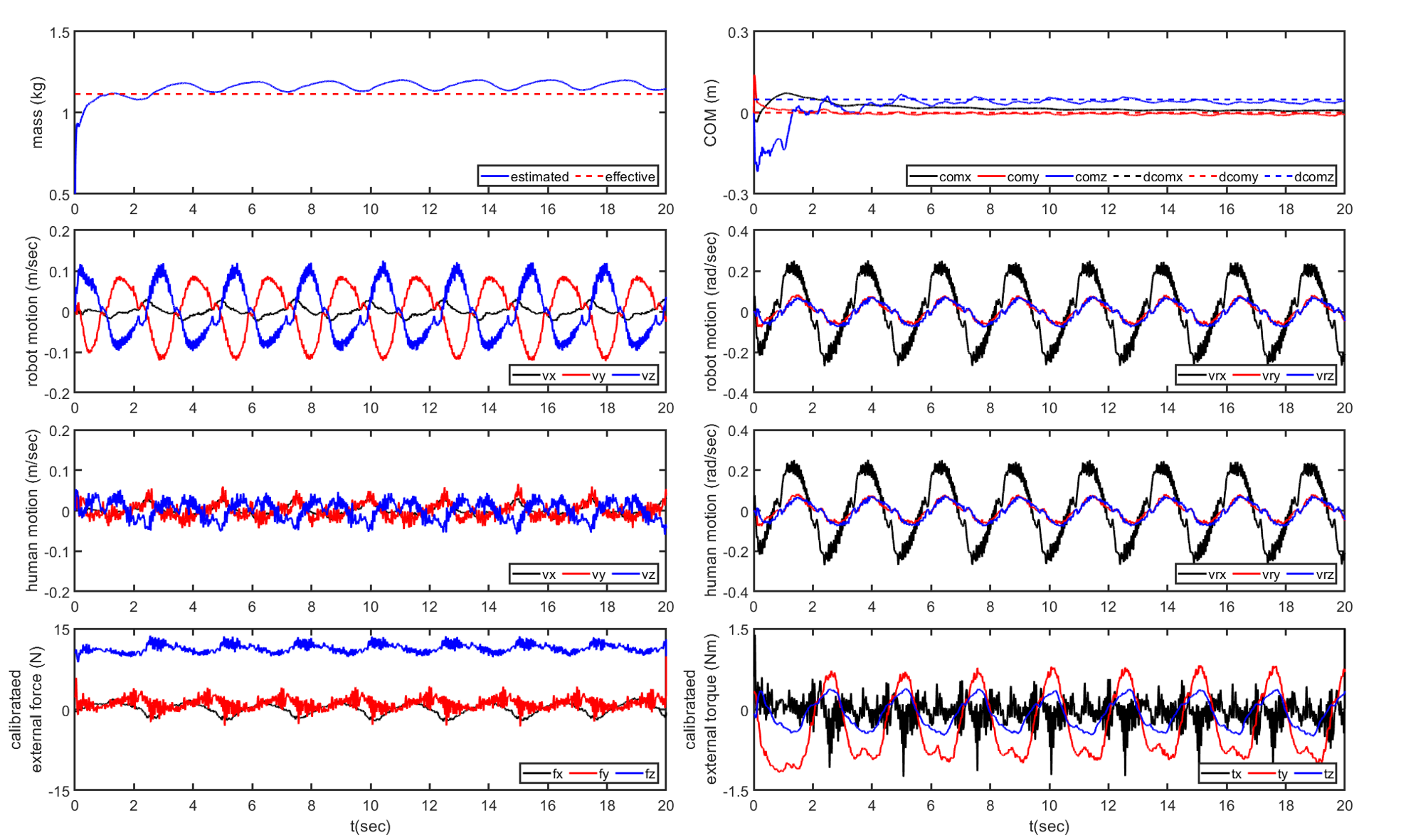}            
    \caption{Object parameter estimation results. The duration of the estimation process is set to 20 sec. The estimation results of the mass and center of mass converge within 10 sec. The robot generates perturbative motion, as depicted by the linear and angular velocity signals of robot end-effector motion, while the linear velocity signal of the human motion remains static. The signals are expressed in the robot end effector frame \{r\}. }
    \label{Fig_4}
\end{figure*}
For parameter estimation, the development of estimation methods such as least squares or adaptive algorithms is crucial. However, to enhance accuracy and precision, the design of a probing signal is also important~\cite{Park}. The object should be perturbed to satisfy the persistent excitation condition, which is achieved by noncolinear angular motion. However, the input perturbation signal should not affect human motion during collaborative tasks to guarantee safety.
Pivoted angular motion with respect to the human grasp point enables the object to be perturbed without significantly affecting the human motion.

Similar to (\ref{eq_4}), the kinematic relation between the robot and hand grasping frames can be represented as follows:
\begin{eqnarray}
	\label{eq_37}
    \begin{array}{l}
    \dot{\boldsymbol{x}}_h = {}^h\boldsymbol{G}^{T}_r \dot{\boldsymbol{x}}_r
    \end{array}
\end{eqnarray}

The pivoted perturbation motion can be realized with reduced kinematics considering only translational motion as follows:
\begin{eqnarray}
	\label{eq_38}
    \begin{array}{l}
    \dot{\boldsymbol{x}}_{h,trans} = {}^h\boldsymbol{G}^{T}_{r,t}  \dot{\boldsymbol{x}}_r
    \end{array}
\end{eqnarray}
where ${}^h\boldsymbol{G}^{T}_{r,t}=\begin{bmatrix} \boldsymbol{I}_3 & |^{h}\boldsymbol{p}_r|_{\times} \end{bmatrix}$. 
The robot end-effector motion in Cartesian space had redundancy in terms of making only translational motion of human grasping frame, thereby allowing the creation of the desired robot motion as follows:
\begin{eqnarray}
	\label{eq_39}
    \begin{array}{l}
    \dot{\boldsymbol{x}}_{r} = {}^h\boldsymbol{G}^{T+}_{r,t} \dot{\boldsymbol{x}}_{h,t}  +  \{ \boldsymbol{I}_6 - {{}^{h}\boldsymbol{G}^{T+}_{r,t}}  {{}^h\boldsymbol{G}^{T}_{r,t}} \}\boldsymbol{z}_{id}
    \end{array}
\end{eqnarray}
where $\boldsymbol{z}_{id} \in \mathbb{R}^{6}$ is an arbitrary vector projected in the null space of ${}^h\boldsymbol{G}^{T}_{r,t}$. By introducing angular perturbation motion to the vector $\boldsymbol{z}_{id}$, collaborative transportation (translational motion) and estimation with object perturbations can be performed simultaneously, with the human grasping point maintained as a pivot~\cite{Cehajic}. In this paper, the angular motions are chosen as $\boldsymbol{z}_{id} = \begin{bmatrix} \boldsymbol{0}_{1 \times 3} &  \boldsymbol{\omega}^{T}_{id} \end{bmatrix} ^ T $, where $ \boldsymbol{\omega}_{id} = -0.2 \cos (2 \pi F_{f} t) \times \boldsymbol{I}_3 $ [rad/s], with $F_f = 0.4$ [Hz].
The object to be estimated is a 1.5 m long aluminum object with additional attached weights, and the true parameters are represented in Table~\ref{tab1}.
The coordinates representing the center of mass are illustrated in Fig.~\ref{Fig_1}; the y-axes of the human and robot frames are aligned with the longitudinal direction of the object, expressed as $^r\boldsymbol{p}_h = [0, -1.5, 0]^T [m]$.
\begin{table}
\begin{center}
\caption{Object parameter estimation results.}
\label{tab1}
\begin{tabular}{| c | c | c | c | c |}
\hline
  & $mass$ [kg] & $com_x$ [mm] & $com_y$ [mm] & $com_z$ [mm]\\
\hline
True & 2.23 & -683 & 0 & 49 \\
\hline
Effective & 1.115 & 0 & 0 & 49 \\ 
\hline
Estimated & 1.131 & 8 & -7 & 44 \\
\hline 
\end{tabular}
\end{center}
\end{table}
In this paper, the object length is assumed to be known before the object is held by the human and robot; well-known artificial intelligence (AI)-based algorithms such as the region-based convolutional neural network (R-CNN) can be used to estimate the object length.

The effective object parameters depend on the human-robot grasping configuration.
In this experiment, the human and the robot grasp both ends of the object as rigidly as possible; additionally, the human participant attempts to hold the object as parallel to the robot as possible. Accordingly, the effective parameters of the object are presented in Table~\ref{tab1}.
Additional sensors are not employed to estimate human information, such as the human wrench or motion.

The objective of this paper is to effectively eliminate the object dynamics during collaborative tasks and not to estimate the true object parameters; thus, the moment of inertia of the object is not estimated because this parameter is less important than the mass and center of mass if motion with large acceleration does not occur during the task.
Moreover, perturbing the object with large accelerations is undesirable because such motion can potentially affect human safety.
In this section, the human and the robot remain static to accurately verify the parameter estimation performance, and in the following section, the estimation is performed during the collaborative transportation task.

The initial values of the estimator are $\hat{\boldsymbol{\theta}}(0)=\boldsymbol{0}_{10}$ and $\boldsymbol{P}(0)=\boldsymbol{I}_{10}$. The weighting matrix of the estimator is set to $\boldsymbol{Q}=10^{-5} \times diag(1.0, 0.01, 1.0, 0.001, 0, 0, 0, 0, 0, 0)$ and $\boldsymbol{R}=diag(200, 1000, 1000, 1000, 1000, 50)$.
The results are represented in Fig.~\ref{Fig_4} and Table~\ref{tab1}.
Since angular perturbation motion $\boldsymbol{z}_{id}$ was realized in the null space of ${}^h\boldsymbol{G}^{T}_{r,t}$, the robot also showed transitional motion.
On the other hand, the hand motion, estimated with (\ref{eq_38}), was well pivoted, as the linear motion remained static, including only some fluctuations, which may be introduced due to the difficulty of humans holding the object consistently during the duration of the estimation.
The angular velocities of the human, robot and object are the same, as they are connected rigidly during the task.

For the estimation, most unmodeled bias and disturbance terms are subtracted from the calibration motion based on the measured external wrench, resulting in the calibrated external wrench depicted in Fig.~\ref{Fig_4}.
The result of $\hat{m}_o$ converges within $t=5 [sec]$ to $\hat{m}_o = 1.131 [kg]$, with a relative error and an effective mass of $0.016 [kg]$.
The results of $^r\hat{\boldsymbol{p}}_o$ converge within $t=8 [sec]$ to $^r\hat{\boldsymbol{p}}_o = [8, -7, 44]^T [mm]$, with a relative error and an effective mass of $[-8, 7,-5]^T [mm]$.
Because the estimation results for the mass and the center of mass both converge within 10 seconds, the required estimation time for our collaborative task scenario is set to 10 seconds.

\subsection{Human-Robot Collaborative Task Results}
A scenario was selected to demonstrate the effectiveness of our proposed methods, including the online parameter estimation strategy and the Cartesian impedance controller considering object dynamics. The scenario involves human-robot collaborative transportation and assembly tasks with an unknown object. A jig, as depicted in Fig.~\ref{Fig_2}, is utilized for the object assembly task. The specific steps of the scenario are as follows:

\begin{figure*}[t]\centering
    \centering
    \includegraphics[scale=0.50]{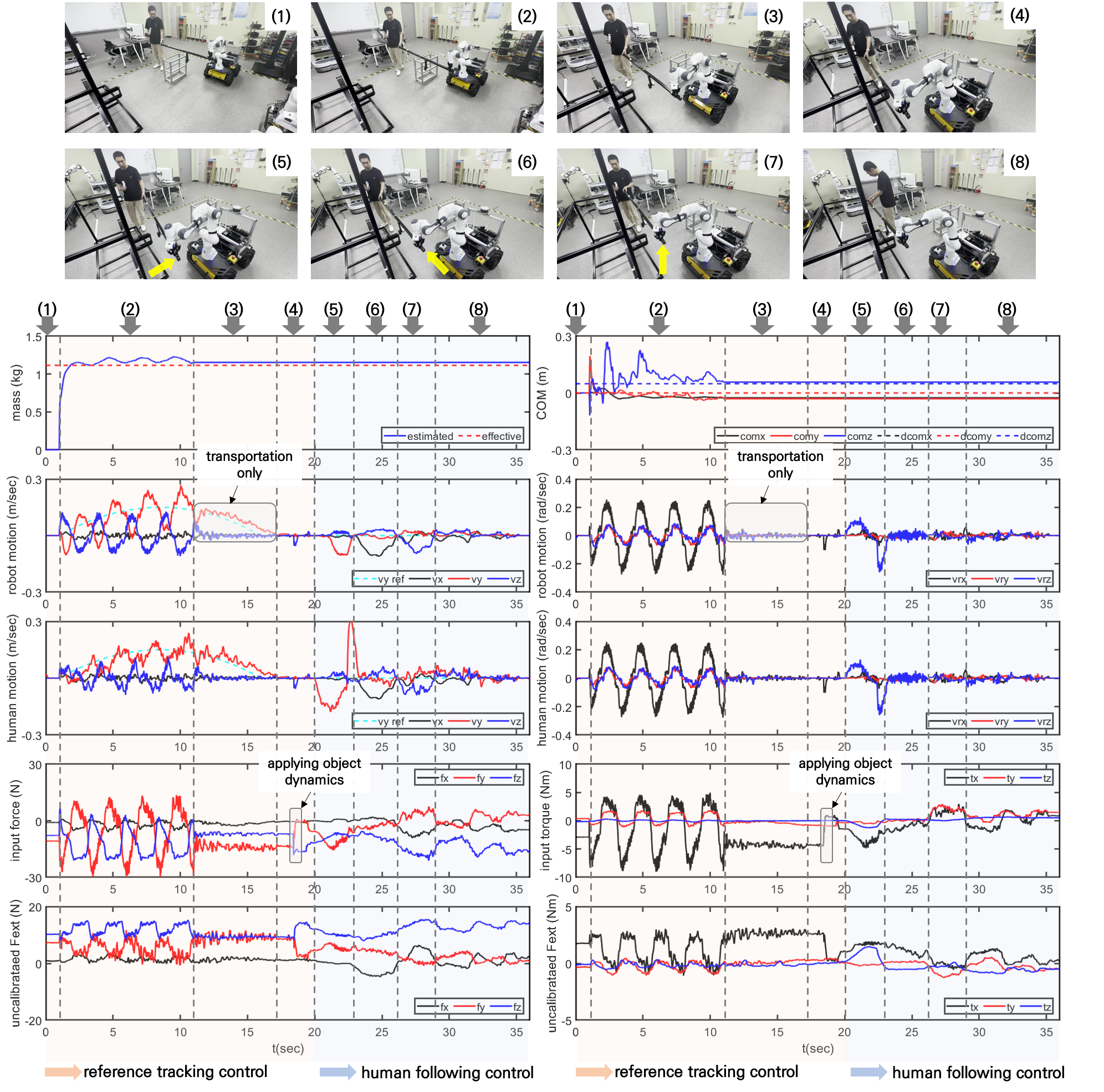}        
    \caption{Demonstration of human-robot collaborative transportation and assembly tasks with the Cartesian object-aware impedance controller. (1) Object lifting; (2) Object parameter estimation during the collaborative transportation task; (3) continuation of the collaborative transportation task; (4) applying estimated object dynamics; the intended human motion following control mainly on the (5) y-axis; (6) x-axis; and (7) z-axis; (8) object assembly onto the jig. The yellow arrows in the snapshots represent the directions of the intended human motion to align the object with the assembly jig. During the transportation task, the mobile manipulator is controlled with a reference tracking control scheme; after compensating for the object dynamics, an object-aware impedance control scheme is applied to follow the intended human motion. The signals are expressed in the robot end effector frame \{r\}.}
    \label{Fig_5}
\end{figure*}

\begin{enumerate}
\item{\textbf{Object lifting:} The human and the robot lift an object together.}
\item{\textbf{Object parameter estimation during the collaborative transportation task:} 
The human and robot transport the object and approach the jig. 
During the transportation task, the parameter estimation is performed while inducing an object perturbation signal in the null space of the desired robot motion.}
\item{\textbf{Continuation of the collaborative transportation task:} Once the estimation process is completed, the perturbation signal is no longer applied, regardless of whether the transportation task was completed. In this scenario, the transportation task continues even after the estimation is completed.}
\item{\textbf{Object dynamics compensation:} After the transportation phase, the object dynamics are compensated for to complete the design of the object-aware impedance controller, as represented in (\ref{eq_34}).}
\item{\textbf{Following the intended human motion:} The human intention following control, also called as follow-me control~\cite{Leonori}, is implemented by setting the stiffness of the desired impedance, represented in (\ref{eq_30}), to zero. This enables active tracking of the intended human motion to locate the end of the object on the robot grasping side at the assembly point of the jig.}
\item{\textbf{Object assembly onto the jig:}
The human accomplishes the assembly task by securely attaching both ends of the object to the jig. Meanwhile, the robot returns to its initial position.}
\end{enumerate}

As reflected in the above scenario, we attempted to minimize the duration of the transportation task by conducting the parameter estimation during this task employing a nonthreatening perturbation signal.
Therefore, by incorporating the object dynamics into the impedance controller in real time, the intended human motion can be transmitted to the robot without any additional action during the scenario.
Note that the human worker performs the task with one hand holding the object, and their other hand is used to manipulate the joystick for the step transition. Therefore, the human can successfully perform the scenario without additional control PCs or coworkers.
The whole scenario is available at \url{https://www.youtube.com/watch?v=bGH6GAFlRgA}, where the task was captured in a single continuous take.

With several practice of collaborative task, it was observed that transmitting intended human motion to the robot through the object becomes more challenging as the longer the object.
The results showed that the use of rotational wrench at the robot side is more efficient to understand the translational human motion.
To incorporate this observation into the weighting matrix of the external wrenches, $\boldsymbol{M}_{\boldsymbol{F}}$, presented in (\ref{eq_34}), we defined $\boldsymbol{P}_d$ as follows:
\begin{eqnarray}
	\label{eq_40}
    \begin{array}{l}     
    \boldsymbol{P}_d = \begin{bmatrix} \begin{bmatrix} 1 & 0 & 0\\ 0 & 0 & 0 \\ 0 & 0 & 0 \end{bmatrix} & \begin{bmatrix} 0 & 0 & 0\\ 0 & 0 & -5 \\ 0 & -4 & 0 \end{bmatrix} \\ 0_3 & 0_3 \end{bmatrix} 
    \end{array}
\end{eqnarray}
With the coordinate system illustrated in Fig.~\ref{Fig_1}, the translational motion of the robot along the +y and +z directions can be adjusted based on the rotational motion of the human along the -z and -y axes. Moreover, the robot's +x transitional motion can be adjusted based on the +x translation motion of the human due to the alignment of the x-axis of the robot's end effector, the human grasping point and the object.

The overall results of the scenario are presented in Fig.~\ref{Fig_5}.
The object parameter estimation results during the nonstationary transportation task are $mass=1.154 [kg]$ and $^{r}\boldsymbol{p}_h=[-0.026, -0.03, 0.057]^T (m)$.
Comparatively, the results exhibited larger errors than those obtained in the stationary task. The bias in the results may be influenced by the human's inability to maintain the alignment between the robot's grasping position and the object during the transportation phase; especially, the negative bias of x and y-axes center of gravity indicates that human participant tends to walk rapidly and more raise the object than the robot based on the coordinate system in Fig.~\ref{Fig_1}.
Nevertheless, the estimation errors are acceptable as the robot could maintain its posture without tilting downward during human intention following control.

During the transportation phase, an impedance controller with nonzero stiffness and damping, $\boldsymbol{K}_e = diag[100, 100, 100, 400, 400, 600]^T$, $\boldsymbol{C}_e = 0.5 \times (2\sqrt{\boldsymbol{K}_e})$, was employed to move the robot's end effector from the initial position, $\boldsymbol{p}_{initial}=[0, 0, 0]^T$, $\boldsymbol{q}_{initial}=[0, 0, 0, 1]^T$, to the final position, $\boldsymbol{p}_{final}=[0, 1.5, 0]^T$, $\boldsymbol{q}_{final}=[0, 0, 0, 1]^T$, by tracking the desired trajectory.
In the whole-body controller defined in (\ref{eq_24}), high-priority tasks (priority 0) include the trajectory tracking control in the Cartesian space and limiting the input torques, while low-priority tasks (priority 1) include maintaining the initial robot arm configuration.
Notably, in the tracking control task, the emphasis lies that the human is allowed to track the desired trajectory, whereas the end effector does not exactly follow the trajectory due to the presence of the identification input.
After the transportation phase is completed, the input force and torques were modulated due to the feedforward incorporation of the object dynamics, as shown in Fig.~\ref{Fig_5}.

To attach the object to the jig, precise alignment is necessary.
To achieve this, the intended human motion following the control strategy was executed by configuring the highest priority task of the whole-body controller to maintain the initial Cartesian position using an impedance controller with zero translational stiffness. The priority 1 task attempts to maintain the initial configuration of the robot arm, as in the previous experiment.
The results show that the external wrench $\boldsymbol{F}_{ext}$ does not significantly increase during the object alignment, indicating that the robot successfully follows the intended human motion. Moreover, the x-, y-, and z-axes of the object position are adjusted by the x-axis of the external force and the z- and y-axes of the external torques, as represented in~(\ref{eq_40}).
Note that some vibrations appear in  $\boldsymbol{F}_{ext}$ during the transportation phase due to the motion of the mobile base, and these vibrations can be eliminated by tuning the velocity controller.

Consequently, the proposed human-robot collaborative method can realize transportation and assembly tasks in a single operation.

\section{Conclusions}

In this paper, an approach for physical human-robot interactions that accounts for unknown object dynamics is proposed, including an online estimation method and a Cartesian impedance controller.
The EKF estimator with an object perturbation input aims to estimate effective object parameters in a specific configuration involving a human, robot, and object.
The object-aware impedance control strategy incorporates the object dynamics in real time, allowing exact transmission of the intended human motion.

Notably, a key feature of this approach is that the following control scheme is realized through physical contact with the object rather than direct contact between the robot and the human.
To achieve this, the direction of the intended human motion is transformed from rotational to linear motion by shaping the desired inertia matrix.
To demonstrate the feasibility of the proposed method in real-world scenarios, transportation and assembly tasks were performed.
The results showed that safe and accurate interactions with long objects can be successfully achieved.

In future works, we will focus on enhancing the parameter estimation performance by effectively removing the uncertainties of human motion and the bias in the external force estimators with physics informed neural network (PINN).


%

\begin{IEEEbiography}[{\includegraphics[width=1in, height=1.25in, clip,keepaspectratio]{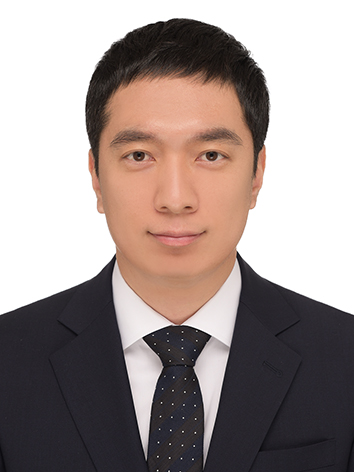}}]
    {Jinseong Park} (Member, IEEE) received B.S., M.S. and Ph.D. degrees in mechanical engineering from the Korea Advanced Institute of Science and Technology (KAIST), Daejeon, Republic of Korea, in 2008, 2010 and 2016, respectively. He was a senior researcher at Hyundai Robotics Co., Ltd., Ulsan, Republic of Korea, from 2016-2017.
    He has been a senior researcher at the Korea Institute of Machinery and Materials (KIMM), Daejeon, Republic of Korea, since 2017. His research interests include optimal control, active vibration control, and fault diagnosis, with an emphasis on mobile manipulators, vehicles, and magnetic levitation systems.
\end{IEEEbiography}

\begin{IEEEbiography}[{\includegraphics[width=1in, height=1.25in, clip,keepaspectratio]{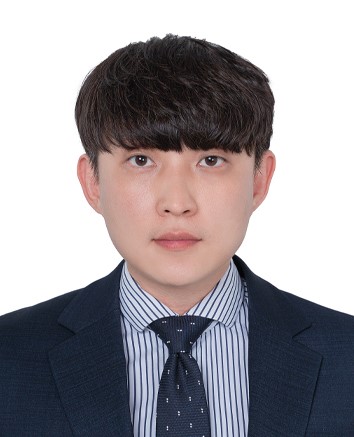}}]
    {Young-Sik Shin} (Member, IEEE) received a B.S. degree in electrical engineering from Inha University, Incheon, South Korea, in 2013, and M.S. and Ph.D. degrees in civil and environmental engineering with a dual degree from the Robotics Program, Korea Advanced Institute of Science and Technology (KAIST), Daejeon, South Korea, in 2015 and 2020, respectively. He is currently a senior researcher with the Korea Institute of Machinery and Materials (KIMM), Daejeon. His research interests include robust simultaneous localization and mapping and navigation using heterogeneous sensors.
\end{IEEEbiography}

\begin{IEEEbiography}[{\includegraphics[width=1in, height=1.25in, clip,keepaspectratio]{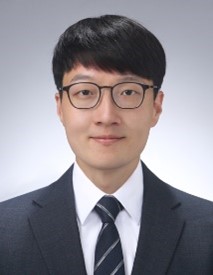}}]
    {Sanghyun Kim} (Member, IEEE) received a B.S. degree in mechanical engineering and a Ph.D. degree in intelligent systems from Seoul National University, South Korea, in 2012 and 2020, respectively. He was a postdoctoral researcher at the University of Edinburgh, U.K., in 2020, and a senior researcher at the Korea Institute of Machinery and Materials (KIMM), South Korea, from 2020 to 2023. He is currently an assistant professor of mechanical engineering at Kyung Hee University. His main research interests include the optimal control of mobile manipulators.
\end{IEEEbiography}

\vfill

\end{document}